\pdfoutput=1

\documentclass[11pt]{article}

\usepackage[]{acl}

\usepackage{times}
\usepackage{latexsym}

\usepackage[T1]{fontenc}

\usepackage[utf8]{inputenc}

\usepackage{ragged2e}
\usepackage{microtype}
\usepackage{pifont}
\usepackage{framed}
\usepackage{times}
\usepackage{soul}
\usepackage{url}
\usepackage[utf8]{inputenc}
\usepackage{graphicx}
\usepackage{amsmath}
\usepackage{amsthm}
\usepackage{booktabs}
\usepackage{algorithm}
\usepackage{algorithmic}
\usepackage{multirow}
\usepackage{subfigure}
\usepackage{amsmath}
\usepackage{diagbox}
\usepackage{color}
\usepackage{tcolorbox}
\usepackage[switch]{lineno}
\graphicspath{{figure/}}

\usepackage{xcolor}

\title{${\rm \textbf{R}^\textbf{2}\textbf{F}}$: A General Retrieval, Reading and Fusion Framework \\ for Document-level Natural Language Inference}

\author{Hao Wang$^{\S}$, Yixin Cao$^{\dag*}$, Yangguang Li$^{\ddag}$, Zhen Huang$^{\S}$ \\
        \bf{Kun Wang$^{\ddag}$, Jing Shao$^{\ddag}$} \\
        $^{\S}$ National University of Defense Technology \\
        $^{\dag}$ Singapore Management University
        $^{\ddag}$ SenseTime}

\begin{document}

\maketitle

\renewcommand{\thefootnote}{\fnsymbol{footnote}}
\footnotetext[1]{Corresponding Author.}
\renewcommand{\thefootnote}{\arabic{footnote}}

\begin{abstract}
Document-level natural language inference (D{\small OC}NLI) is a new challenging task in natural language processing, aiming at judging the entailment relationship between a pair of hypothesis and premise documents.
Current datasets and baselines largely follow sentence-level settings, but fail to address the issues raised by longer documents. 
In this paper, we establish a general solution, named Retrieval, Reading and Fusion (R$^2$F) framework, and a new setting, by analyzing the main challenges of D{\small OC}NLI: interpretability, long-range dependency, and cross-sentence inference. 
The basic idea of the framework is to simplify document-level task into a set of sentence-level tasks, and improve both performance and interpretability with the power of evidence.
For each hypothesis sentence, the framework retrieves evidence sentences from the premise, and reads to estimate its credibility.
Then the sentence-level results are fused to judge the relationship between the documents.
For the setting, we contribute complementary evidence and entailment label annotation on hypothesis sentences, for interpretability study.
Our experimental results show that R$^2$F framework can obtain state-of-the-art performance and is robust for diverse evidence retrieval methods. 
Moreover, it can give more interpretable prediction results.
Our model and code are released at \url{https://github.com/phoenixsecularbird/R2F}.

\end{abstract}

\section{Introduction}
\label{introduction}

Natural Language Inference (NLI) is the task of determining whether a \textit{hypothesis} is entailed or not in a \textit{premise}. 
While earlier works~\cite{bowman:snli,samuel:mnli,wang:qnli,nie:anli} assume that both hypothesis and premise are single sentences, recent research pays more attention on document-level task, namely Document-level NLI (D{\small OC}NLI) \cite{yin:docnli}.
The task can enlarge the task scope to judge the variability of semantic expression for many Natural Language Processing (NLP) tasks, e.g., exposure bias~\cite{bengio:bias,arora:bias} alleviation for text summarization~\cite{evan:summary,shay:summary}, and human-manipulated news articles recognition for automatic fake news detection~\cite{laks:detail,ji:real}.

\begin{figure*}[!h]
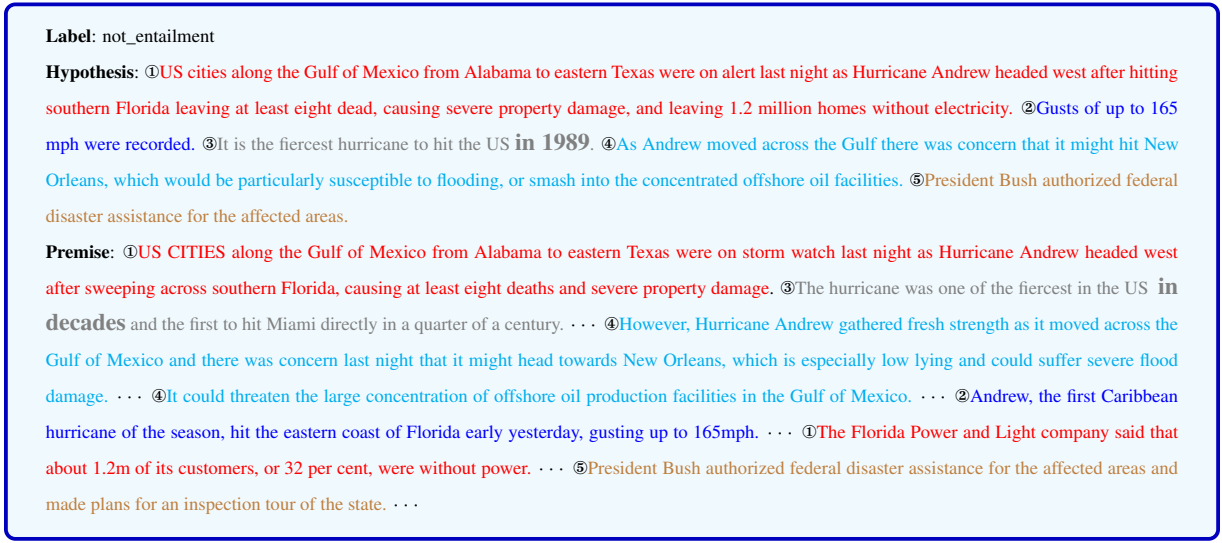

\begin{tcolorbox}[colback=cyan!5!white, colframe=blue!75!black]

{\scriptsize \textbf{Label}: not\_entailment}

{\scriptsize \textbf{Hypothesis}:  \ding{172}\textcolor{red}{US cities along the Gulf of Mexico from Alabama to eastern Texas were on alert last night as Hurricane Andrew headed west after hitting southern Florida leaving at least eight dead, causing severe property damage, and leaving 1.2 million homes without electricity.} \ding{173}\textcolor{blue}{Gusts of up to 165 mph were recorded.} \ding{174}\textcolor{gray}{It is the fiercest hurricane to hit the US {\small \textbf{in 1989}}.} \ding{175}\textcolor{cyan}{As Andrew moved across the Gulf there was concern that it might hit New Orleans, which would be particularly susceptible to flooding, or smash into the concentrated offshore oil facilities.} \ding{176}\textcolor{brown}{President Bush authorized federal disaster assistance for the affected areas.}}

{\scriptsize \textbf{Premise}: \ding{172}\textcolor{red}{US CITIES along the Gulf of Mexico from Alabama to eastern Texas were on storm watch last night as Hurricane Andrew headed west after sweeping across southern Florida, causing at least eight deaths and severe property damage}. \ding{174}\textcolor{gray}{The hurricane was one of the fiercest in the US { \small \textbf{in decades}} and the first to hit Miami directly in a quarter of a century.} $\cdots$ \ding{175}\textcolor{cyan}{However, Hurricane Andrew gathered fresh strength as it moved across the Gulf of Mexico and there was concern last night that it might head towards New Orleans, which is especially low lying and could suffer severe flood damage.} $\cdots$ \ding{175}\textcolor{cyan}{It could threaten the large concentration of offshore oil production facilities in the Gulf of Mexico.} $\cdots$ \ding{173}\textcolor{blue}{Andrew, the first Caribbean hurricane of the season, hit the eastern coast of Florida early yesterday, gusting up to 165mph.} $\cdots$ \ding{172}\textcolor{red}{The Florida Power and Light company said that about 1.2m of its customers, or 32 per cent, were without power.} $\cdots$ \ding{176}\textcolor{brown}{President Bush authorized federal disaster assistance for the affected areas and made plans for an inspection tour of the state.} $\cdots$}

\end{tcolorbox}
\caption{A sample of D{\small OC}NLI dataset. For each sample, only the entailment label between the documents is annotated. For display, we mark each hypothesis sentence and its corresponding premise sentences (namely the evidences, not annotated in the original dataset) with the same number and color. The sample is annotated as \textit{not entailment} due to the disinformation of ``\textit{in 1989}'' in the third hypothesis sentence. The premise is partly omitted.}
\label{sample}
\end{figure*}

Compared with sentence-level NLI, D{\small OC}NLI poses many new challenges, while there are only a few datasets and models. 
In terms of datasets, \citet{yin:docnli} reformat some mainstream NLP tasks, e.g., text summarization and question answering, and build the \textit{first} large scale dataset D{\small OC}NLI with over 1 million document pairs\footnote{\citet{yin:docnli} propose the task and annotate the dataset with the same name D{\tiny OC}NLI.}.
However, the dataset does not provide evidence annotation about how the labels are inferred, i.e., which hypothesis sentences lead to semantic inconsistency, or which premise sentences help to decide the entailment relationship.
As shown in Figure~\ref{sample}, although the sample is annotated as \textit{not entailment}, most of the hypothesis sentences are actually entailed. 
By contrast, the detailed disinformation of ``\textit{in 1989}'' in the third hypothesis sentence eventually decides the entailment relationship between the documents. 
For each  hypothesis sentence, only several premise sentences are enough to serve as the exact evidence to judge its own sentence-level entailment label.

In this paper, we argue that evidence discovery is important and challenging for D{\small OC}NLI. 
Our pilot experiments in Section \ref{influence of evidence retrieval} and \ref{interpretability study} show that randomly selected evidences can still contribute to comparable performance. 
Thus, only the black-box models may be not so convincing.
However, to annotate evidence for evaluation is non-trivial. 
For each hypothesis sentence, on one hand, it may refer to multiple premise sentences. 
On the other hand, there may be several evidence groups, where each group can independently serve the label prediction. 
We highlight this as \textbf{interpretability} challenge.

In term of models, current baselines~\cite{yin:docnli,qiu:match} still largely follow sentence-level NLI.
They either concatenate two documents for mutual information interaction for classification, or encode them separately for semantic match with document-level representations.
However, except for the interpretability issue, they will leave the following challenges  unexplored:

$\bullet$~\textbf{Long-range Dependency Modeling}~The task requests to handle a pair of long documents at the same time, where we observe that 29.81\% samples of D{\small OC}NLI dataset~\cite{yin:docnli} contain more than 500 \textit{words}\footnote{A word may correspond to multiple tokens for PLMs.}, while 10.47\% samples contain more than 1000 \textit{words}.
This will not only exceed the input limit of Pre-trained Language Models (PLMs), but also make it far more difficult to capture long-range dependency.
Necessary information interaction between the hypothesis and some key premise sentences may not be guaranteed.
Besides, most contexts are uninformative for entailment inference and will only serve as noise. 

$\bullet$~\textbf{Cross-sentence Inference}~To judge the relationship between the documents, it is supposed to consider all hypothesis sentences, where the detailed disinformation issue still remains unsolved.
Besides, the verification of one hypothesis sentence may request to combine multiple and distant premise sentences, different from the sentence pair mode in sentence-level NLI.
As shown in Figure~\ref{sample}, to process the first one, it needs to take both the first and sixth premise sentences (all in red fonts).

\begin{figure*}[!h]
\centerline{\includegraphics[scale=0.38]{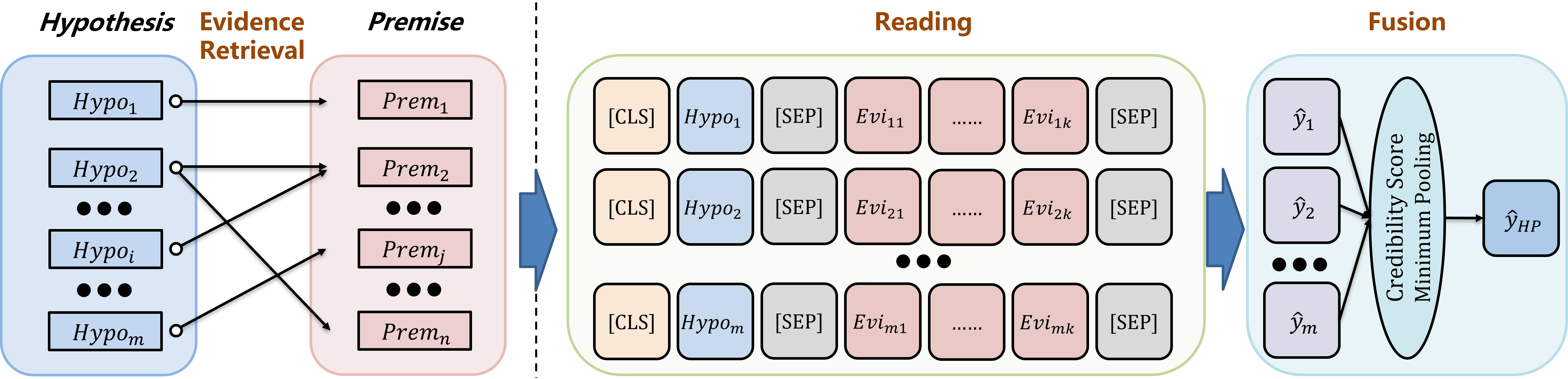}}
\caption{$\rm{R^2F}$ framework. For each hypothesis sentence, the framework firstly \textbf{retrieves} evidence sentences from the premise, and then \textbf{reads} to estimate the credibility upon the evidences. Finally it \textbf{fuses} the sentence-level results, and judges the entailment relationship between these two documents. $\hat{y}_i$ and $\hat{y}_{HP}$ are the credibility score of the \textit{i}-th hypothesis sentence and the sample, while $Evi_{ij}$ is the \textit{j}-th evidence of the \textit{i}-th hypothesis sentence.}
\label{model}
\end{figure*}

In this paper, we establish a general solution, named Retrieval, Reading and Fusion (R$^2$F) framework, and a new setting for the task.
The basic idea of the framework is to simplify document-level classification task into a set of sentence-level tasks, and then improve both performance and interpretability with the power of evidence.
Specifically, the framework splits the hypothesis document into sentences.
Then for each hypothesis sentence, it retrieves evidence sentences from the premise, and reads to estimate its credibility score upon the evidences.
Finally, it fuses the sentence-level results and judge the entailment relationship between the two documents. 
For the setting, we contribute complementary fine-grained annotations for interpretability study.
For each hypothesis sentence, we manually annotate entailment label and several evidence groups, where each group is enough to independently infer the label.
In summary, our contributions are as follows:

$\bullet$~We propose a Retrieval, Read and Fusion framework as a general solution for D{\small OC}NLI task.

$\bullet$~We contribute complementary evidence and entailment label annotation for each hypothesis sentence on a subset of D{\small OC}NLI dataset for interpretability study.
 
$\bullet$~Our experimental results on D{\small OC}NLI dataset indicate that the framework obtains state-of-the-art performance.
Besides, it is robust for diverse retrieval methods.
Moreover, the framework can give more interpretable prediction results.

\section{${\rm \textbf{R}^\textbf{2}\textbf{F}}$ Framework}

Our $\rm{R^2F}$ framework aims at a general solution for D{\small OC}NLI task with interpretability, i.e., to obtain corresponding evidence and predict entailment label for each hypothesis sentence.
As shown in Figure~\ref{model}, the framework consists of 3 components, namely evidence retrieval, reading for credibility estimation, and credibility fusion. 
For efficiency, the retrieval component is an independent unit to provide evidence input for the other two components, which are optimized jointly.

\subsection{Task Formulation}
Similar to previous sentence-level NLI tasks, for each sample in D{\small OC}NLI task, given a hypothesis document \textit{H} and a premise document \textit{P}, it is requested to judge the entailment relationship \textit{R} between these two documents. 
Here, \textit{R} $\in$ \{``\textit{entailment}'', ``\textit{not\_entailment}''\} for D{\small OC}NLI dataset, but may not be restricted to binary classification.

\subsection{Evidence Retrieval}
\label{evidence retrieval}

Given each sample, we split the hypothesis into sentences and retrieve evidence sentences from the premise.
Formally, we split the hypothesis \textit{H} and the premise \textit{P} into single sentences \{$H_1$, $H_2$, $\cdots$, $H_m$\} and \{$P_1$, $P_2$, $\cdots$, $P_n$\}, through  NLTK tool\footnote{\url{https://www.nltk.org/}}.
Here, \textit{m} and \textit{n} are the sentence numbers.

For each hypothesis sentence $H_i$,  we respectively utilize the following retrieval methods to calculate the \textit{relevance score} with all premise sentences. 
Then according to the scores, we remain top \textit{K} sentences as the corresponding evidence.
The value of \textit{K} is a trade-off between evidence recall and precision.
A lower value pursues higher evidence precision, but may lead to evidence missing, while a higher value guarantees higher evidence recall, but may introduce too many uninformative sentences as noise.
Moreover, to keep and utilize contextual information, for each hypothesis sentence, we reorder the evidence sentences according to their original order in the premise.

To calculate the relevance score, we take several sparse and dense retrieval methods into consideration:

$\bullet$~\textbf{ROUGE-1}~Inspired by~\citet{mao:rouge1} and~\citet{zhang:rouge2}, we adopt ROUGE-1 retrieval.
For a pair of sentences, this sparse retrieval method focuses on \textit{n-gram match} of the pair to calculate ROUGE-1 score as the relevance metric.
We take it as the \textbf{main retrieval method}.

$\bullet$~\textbf{BM25}\footnote{We adopt the implementation from \url{https://github.com/dorianbrown/rank_bm25}.}~BM25 is one of the most advanced sparse retrieval methods. 
We take all premise sentences as the corpus.
For a pair of sentences, BM25 involves not only the pair itself but also the whole corpus, to count \textit{term frequency} and \textit{inverse-document frequency} to obtain the relevance score.

$\bullet$~\textbf{SimCSE}\footnote{We adopt unsupervised and supervised version of RoBERTa$_{base}$.}~Inspired by \citet{gao:simcse}, we utilize SimCSE \cite{gao:simcse}, a strong sentence embedding model, as dense retrieval method for \textit{semantic match}.
For a pair of sentences, we take the cosine similarity of the sentence embeddings as the relevance score.

Except for above retrieval methods, we also adopt another simple but effective strategy.
If a hypothesis sentence is a \textit{substring} of the premise, then it is naturally entailed in the premise through \textit{string match} and does not need further study.

\subsection{Reading \& Credibility Estimation}

For each hypothesis sentence, we concentrate informative premise sentences and filter out most noisy ones through evidence retrieval.
Then this component aims to estimate its credibility, which involves the hypothesis sentence itself and several evidence sentences.
This is different from conventional sentence-level NLI, which studies the relationship between a pair of sentences.
Hence we adopt reading models.
Our general $\rm{R^2F}$ framework is compatible to any arbitrary reading models, which may be enhanced by several advanced learning technologies, e.g., graph neural network~\cite{kipf:gcn,petar:gat}, commonsense knowledge injection~\cite{zhang:erine,wang:kadapter}, and syntactic structure analysis~\cite{lu:parsing}.
Herein, we adopt a simple and straightforward one  without loss of generality.

Specifically, for a hypothesis sentence $H_i$, and its corresponding evidence \{$Evi_{i1}$, $Evi_{i2}$, $\cdots$, $Evi_{iK}$\}, we concatenate them all as the input:
\begin{equation}
[CLS]\;H_i\;[SEP]\;Evi_{i1}\;Evi_{i2}\;\cdots\;Evi_{iK}\;[SEP]
\end{equation}
Then we leverage transformer-based pre-trained language model, i.e., RoBERTa~\cite{liu:roberta}, to encode the input sequence.
Through the multi-head self-attention mechanism~\cite{ashish:attention}, token-level information interaction among the hypothesis sentence and all its evidences is conducted.
Besides, since evidence are concentrated, it is much easier to handle the multiple evidence combination issue for cross-sentence inference.

Then the credibility score is calculated through a Multi Layer Perceptron (MLP) with \textit{sigmoid} activation function:
\begin{equation}
\hat{y}_i = {\rm Sigmoid}({\rm MLP}(h_{i}))
\end{equation}
where $h_{i}$  is the hidden state of the special [CLS] token, and is taken as the inference vector of  $H_i$.
Besides, $\hat{y}_i$ $\in$ [0.0, 1.0] is the credibility score of $H_i$, and a higher score means that the sentence is more likely to be entailed by the premise.

\subsection{Credibility Fusion}

After reading, the inference vector $h_{i}$ and credibility score $\hat{y}_i$ of each hypothesis sentence $H_i$ is obtained.
Nevertheless, the reading model cannot be trained directly  since the detailed entailment label of $H_i$ is not available.
To this end, we fuse the sentence-level results to judge the entailment relationship between the documents, and indirectly train the model through document-level entailment label.
Besides, the fusion process is also expected to solve the detailed disinformation issue for cross-sentence inference, and expand the interpretability of the framework.
Herein, we design three fusion methods for comparison.

$\bullet$~\textbf{Credibility Score Minimum Pooling}~The logic basis for this method is that if the premise entails the hypothesis, then it will entail all the hypothesis sentences, even the one with the lowest credibility score.
By contrast, if the premise does not entail the hypothesis, then it will conflict to at least one hypothesis sentence.
This one is expected to be assigned with the lowest credibility score.

Formally, for a pair of documents \textit{H} and \textit{P}, the credibility scores of the hypothesis sentences are \{$\hat{y}_1$, $\hat{y}_2$, $\cdots$, $\hat{y}_m$\}.
Then the credibility score of the sample is calculated as:
\begin{equation}
\hat{y}_{HP} = min(\{\hat{y}_1, \hat{y}_2, \cdots, \hat{y}_m\})
\end{equation}

For this fusion method, the sample prediction result comes from that of the  least credible hypothesis sentence, i.e., with the lowest credibility score.
The framework is requested to conduct internal contrast among the hypothesis sentences to decide the least credible one. 
Thus, it is forced to understand the entailment relationship between each hypothesis sentence and its corresponding evidences although without direct entailment label.
During prediction, the credibility score $\hat{y}_i$ is utilized to predict the entailment label of hypothesis sentence $H_i$.
We take this as the \textbf{main fusion method}.

$\bullet$~\textbf{Inference Vector Minimum Pooling}~For this method, we conduct minimum pooling on the inference vector $h_{i}$ rather than the credibility score $\hat{y}_i$.
For the inference vector of the sample $h_{HP}$, the \textit{j}-th dimension is:
\begin{equation}
h^j_{HP} = min(\{h^j_1, h^j_2, \cdots, h^j_m\})
\end{equation}
Then the credibility score of the sample is:
\begin{equation}
\hat{y}_{HP} = {\rm Sigmoid}({\rm MLP}(h_{HP}))
\end{equation}

Many conventional neural models tend to adopt this fusion method for better performance, but suffer from low interpretability, since the practical meaning of each dimension of the inference vector can hardly be probed.

$\bullet$~\textbf{Gaussian Kernel Pooling}~To further explore the influence of fusion methods, we adopt Gaussian Kernel Pooling~\cite{xiong:kernel,liu:kernel,sheng:zoom}.
Specifically, we utilize \textit{C} Gaussian kernels $\{K_j\}_{j=1}^{C}$.
For a credibility score $\hat{y}_i$, the output of the \textit{j}-th kernel is:
\begin{equation}
V_i^j=\exp ({-\frac {(\hat{y}_i-\mu_j)^2} {2{\sigma}^2_j}})
\end{equation}
where $\mu_j$ and $\sigma_j$ are respectively the mean and width of the \textit{j}-th kernel.
In this way, the score $\hat{y}_i$ is projected to a kernel vector $V_i$ $\in$ $\mathcal{R}^C$.
The kernel vector of the sample is:
\begin{equation}
V_{HP} = \frac {1} {m} \sum_{i=1}^{m} V_i \in \mathcal{R}^C
\label{eq}
\end{equation}
And the credibility score of the sample is:
\begin{equation}
\hat{y}_{HP} = {\rm Sigmoid}({\rm MLP}(V_{HP}))
\end{equation}

This is a mean pooling method, which conducts mean pooling on the corresponding kernel vectors, rather than the credibility scores.

During training, the loss function of the sample is set as binary cross entropy loss:
\begin{equation}
\mathcal{L} = y_{HP}log(\hat{y}_{HP}) + (1-y_{HP})log(1-\hat{y}_{HP})
\end{equation}
where $y_{HP}$ is the sample entailment label, 1 for \textit{entailment} samples while 0 for \textit{not entailment} ones.
During prediction, we set a threshold on the sample credibility score $\hat{y}_{HP}$ to obtain the result.

\section{Experiment}

\subsection{Dataset}

We conduct our experiments on D{\small OC}NLI dataset~\cite{yin:docnli}, which is a newly proposed and the only large scale dataset in the field.
The detailed statistic information is shown in Table \ref{dataset}.
The training set comes with balanced label distribution, while the development and test sets come with pretty unbalanced ones.

\begin{table}[!h]
\setlength\tabcolsep{3pt}
\centering
\begin{tabular}{cccc}
\hline
\textbf{Set} & \textbf{Entailment} & \textbf{Not\_Entailment} & \textbf{Total} \\ \hline
train        & 466,653             & 475,661                  & 942,314        \\
dev          & 28,890              & 205,368                  & 234,258        \\
test         & 33,128              & 233,958                  & 267,086        \\ \hline
\end{tabular}
\caption{Statistic information of D{\small OC}NLI dataset.}
\label{dataset}
\end{table}

\subsection{Complementary Annotation}

For interpretability study, we contribute complementary annotation\footnote{The annotation is also released.}.
The hypothesis sentences may involve cross-sentence inference and request multiple evidences.
Besides, they may also correspond to several evidence groups, where each group itself is enough to independently infer the entailment label.
Thus the annotation process needs heavy workload and comes with great complexity.
Herein, we adopt a \textit{proposal and correction} annotation strategy.
Specifically, for each hypothesis sentence, we firstly retrieve candidate evidences through the diverse methods in Section \ref{evidence retrieval}.
Then we manually check, remove repeated or unrelated ones and add missing ones, and combine several evidence groups.
Finally, we decide the entailment label according to the evidences.

\begin{figure}[!h]
\begin{tcolorbox}[colback=cyan!5!white, colframe=blue!75!black]

{\scriptsize \textbf{Label}: entailment}

{\scriptsize \textbf{Hypothesis Sentence}:  Tony Abbott will withdraw Australia's ambassador to Indonesia.}

{\scriptsize \textbf{Evidence Group 1}: Prime Minister Tony Abbott said Australia will withdraw its ambassador to Indonesia in an unprecedented diplomatic response to the executions of Myuran Sukumaran and Andrew Chan.}

{\scriptsize \textbf{Evidence Group 2}: Mr Prasetyo shrugged off diplomatic backlash from Australia after Prime Minister Tony Abbott slammed the executions as ``cruel and unnecessary'' and announced he would withdraw Australia's ambassador to Indonesia Paul Grigson.}

\end{tcolorbox}
\caption{Annotation example for hypothesis sentence. The hypothesis sentence is annotated as \textit{entailment} with two evidence groups.}
\label{annotation exaple}
\end{figure}

Due to the heavy workload and great complexity, we manually annotate 100 longer samples (all over 800 words) randomly selected from the \textit{test} set, which contain more than 350 hypothesis sentences in total.
An annotation example is shown in Figure~\ref{annotation exaple}.
The hypothesis sentence is annotated as \textit{entailment} with two evidence groups.
More detailed statistic information is summarized in Appendix~\ref{annotation example}.

\begin{table*}[h]
\centering
\setlength\tabcolsep{4pt}
\begin{tabular}{ccccccccc}
\hline
\multirow{2}{*}{\textbf{Model}} & \multicolumn{3}{c}{\textbf{Entailment}}          & \multicolumn{3}{c}{\textbf{Not\_Entailment}}     & \multicolumn{2}{c}{\textbf{Total}}    \\
                                & \textbf{P}     & \textbf{R}     & \textbf{F1}    & \textbf{P}     & \textbf{R}     & \textbf{F1}    & \textbf{Micro F1} & \textbf{Macro F1} \\ \hline
Concatenation-Longformer$_{base}\clubsuit$       & -              & -              & 44.42          & -              & -              & -              & -                 & -                 \\
Concatenation-Longformer$_{base}\spadesuit$      & 29.73          & 80.32          & 43.39          & 96.33          & 73.11          & 83.13          & 74.01             & 63.26             \\
Concatenation-RoBERTa$_{base}\spadesuit$         & 39.73          & \textbf{91.57} & 55.41          & \textbf{98.54} & 80.33          & 88.51          & 81.72             & 71.96             \\
Semantic Match-RoBERTa$_{base}\triangle$         & 31.60          & 85.49          & 46.15          & 97.29          & 73.80          & 83.93          & 75.25             & 65.04             \\
\textit{ours} $\rm R^2F$-RoBERTa$_{base}$        & \textbf{43.52} & 87.04          & \textbf{58.02} & 97.86          & \textbf{84.01} & \textbf{90.41} & \textbf{84.38}    & \textbf{74.22}    \\ \hline \hline
Concatenation-RoBERTa$_{large}\clubsuit$         & 45.15          & \textbf{94.99} & 61.21          & \textbf{99.16} & 83.66          & 90.75          & 85.06             & 75.98             \\
Concatenation-RoBERTa$_{large}\spadesuit$        & 44.70          & 94.74          & 60.74          & 99.11          & 83.40          & 90.58          & 84.81             & 75.66             \\
\textit{ours} $\rm R^2F$-RoBERTa$_{large}$       & \textbf{50.61} & 88.07          & \textbf{64.28} & 98.11          & \textbf{87.82} & \textbf{92.68} & \textbf{87.86}    & \textbf{78.48}    \\ \hline
\end{tabular}
\caption{Model performance on the \textit{test} set. $\clubsuit$ and $\spadesuit$ denote the \textit{original} and \textit{reproduced} results of the model from~\citet{yin:docnli}. $\triangle$ denotes the \textit{reproduced} results of the model modified from~\citet{qiu:match}.}
\label{test results}
\end{table*}

\subsection{Evaluation Metric}

For D{\small OC}NLI task, we adopt micro and macro F1 scores, and attach more importance to the latter.
Due to the unbalanced label distribution, even majority guess model will obtain high micro F1 score, but pretty low macro F1 score. 
For sentence-level evaluation, we adopt evidence precision, recall and F1 for retrieval, while micro and macro F1 scores for label prediction.
Moreover, inspired by~\citet{thorne:fever}, we propose a more strict metric \textit{full accuracy}. 
Herein, evidence recall requests to find at least one complete evidence group, and full accuracy further requests correct label prediction.

\subsection{Experiment Setup}

Our $\rm{R^2F}$ framework is implemented through Pytorch 1.8.0. 
We adopt AdamW optimizer, keep a random number seed of 42, set max input length as 256, and set mini batch size as 8 with gradient accumulation step as 4.
For base encoder, we choose initial learning rate of 1e-5, while for large encoder, we choose 5e-6.
For evidence retrieval, we set \textit{K} as 5.
During prediction, we adopt a threshold of 0.5.
More setup is shown in Appendix~\ref{setup}.

\subsection{Baseline}

Since D{\small OC}NLI is still a new task, we adopt the concatenation model from~\citet{yin:docnli}, and modify the semantic match model from~\citet{qiu:match} for comparison.
Please refer to the original papers and Appendix~\ref{baseline} for detailed information.

\section{Results}

\subsection{Main Results}

Main results of our framework on the test set are displayed in Table~\ref{test results}.
The framework obtains the best performance with the highest micro and macro F1 scores, with both base and large encoders, indicating its strength.
The situation is similar on the development set in Appendix~\ref{performance on dev set}.

All models show far better performance (higher F1 score) on \textit{not entailment} samples than \textit{entailment} samples, while \textit{entailment} samples still come with even higher recall.
This may be due to the label distribution difference among different sets (in Table~\ref{dataset}).
Among the baselines, semantic match model conducts coarse-grained information interaction between the hypothesis and premise through document-level vector representations.
Thus it shows pretty low performance although it can avoid possible key information missing caused by the truncation of overlong samples.
This indicates that fine-grained information interaction is essential for this task.
Besides, concatenation model with Longformer encoder, although it is able to handle much longer inputs, shows much lower performance than that with RoBERTa encoder, which truncates the inputs.
The simplified multi-head self-attention mechanism in Longformer encoder~\cite{iz:longformer} seems not competent for fine-grained information interaction in the task.

\subsection{Performance vs Sample Length}

\begin{figure}[h]
\centering
\centerline{\includegraphics[scale=0.49]{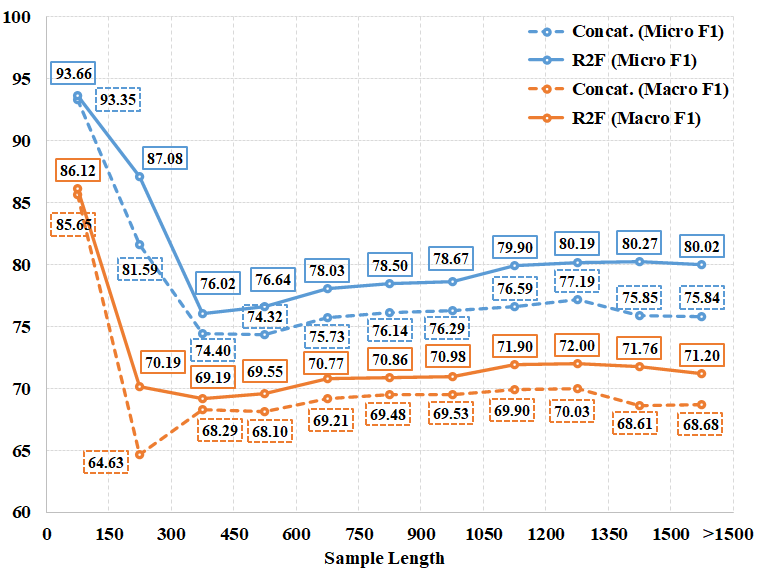}}
\caption{Model performance with varying sample length on the \textit{test} set. Concatenation baseline and R$^2$F framework with RoBERTa$_{base}$ encoder are compared. The horizontal axis is the sample length in the number of words. The vertical axis is the performance.}
\label{test length}
\end{figure}

To examine the ability of our framework on processing long inputs, the performance with varying sample length on the test set is shown in Figure~\ref{test length}.  
Here, the length of a sample is counted in the number of words.
Both models obtain far better performance on shorter samples than longer ones (more than 10\% absolute difference on both micro F1 and macro F1 scores).
Hence it is still a  relatively difficult problem to process longer samples.
Moreover, our R$^2$F framework consistently and greatly outperforms the strongest concatenation baseline on samples with varying length.
Especially, it shows much higher performance on longer samples.
These indicate the framework is able to handle longer ones efficiently through breaking the document-level task into sentence-level task with the retrieval, reading and fusion process.
On the development set, the tendency is similar, and detailed results are displayed in Appendix~\ref{performance on dev set}.

\subsection{Influence of Evidence Retrieval}
\label{influence of evidence retrieval}

To investigate the influence of evidence retrieval, we focus on different retrieval methods, and take those mentioned in Section~\ref{evidence retrieval} for comparison.
Besides, we also adopt a random baseline, which adopts \textit{K} randomly selected premise sentences as the evidences.

\begin{table}[!h]
\centering
\setlength\tabcolsep{4pt}
\begin{tabular}{ccccc}
\hline
\multirow{2}{*}{\textbf{\begin{tabular}[c]{@{}c@{}}Retrieval\\  Method\end{tabular}}} & \multicolumn{2}{c}{\textbf{Dev Set}} & \multicolumn{2}{c}{\textbf{Test Set}} \\
                                                                                      & \textbf{Mi F1}    & \textbf{Ma F1}   & \textbf{Mi F1}    & \textbf{Ma F1}    \\ \hline
Random                                                                                & 83.27             & 71.77            & 82.45             & 70.98             \\
ROUGE-1                                                                               & 85.58             & 75.44            & 84.38             & 74.22             \\
BM25                                                                                  & 85.55             & 75.36            & 83.98             & 73.71             \\
SimCSE$^*$                                                                            & \textbf{86.13}    & \textbf{75.57}   & \textbf{85.34}    & \textbf{74.74}    \\
SimCSE$^\#$                                                                           & 84.87             & 74.33            & 83.65             & 73.28             \\ \hline
\end{tabular}
\caption{D{\small OC}NLI performance with different evidence retrieval methods. $^*$ and $^\#$ respectively denote \textit{unsupervised} and \textit{supervised} version, and the same below.}
\label{retrieval}
\end{table}

As shown in Table~\ref{retrieval}, the random baseline can still obtain relatively high performance although it cannot outperform the concatenation baseline.
This may be due to the \textit{evidence dependency bias} discussed in Section~\ref{interpretability study}.
Besides, for shorter samples with only a few premise sentences, evidence retrieval will show less importance.
All other retrieval methods can contribute to higher performance than the concatenation baseline, indicating the strength of the framework on the task and its robustness for diverse retrieval methods.
However, supervised SimCSE, although trained on human-annotated NLI benchmarks, shows the lowest performance, which may suffer transfer issue caused by the domain difference between its own training data and D{\small OC}NLI dataset.

\subsection{Influence of Fusion Method}

\begin{figure}[h]
\centerline{\includegraphics[scale=0.26]{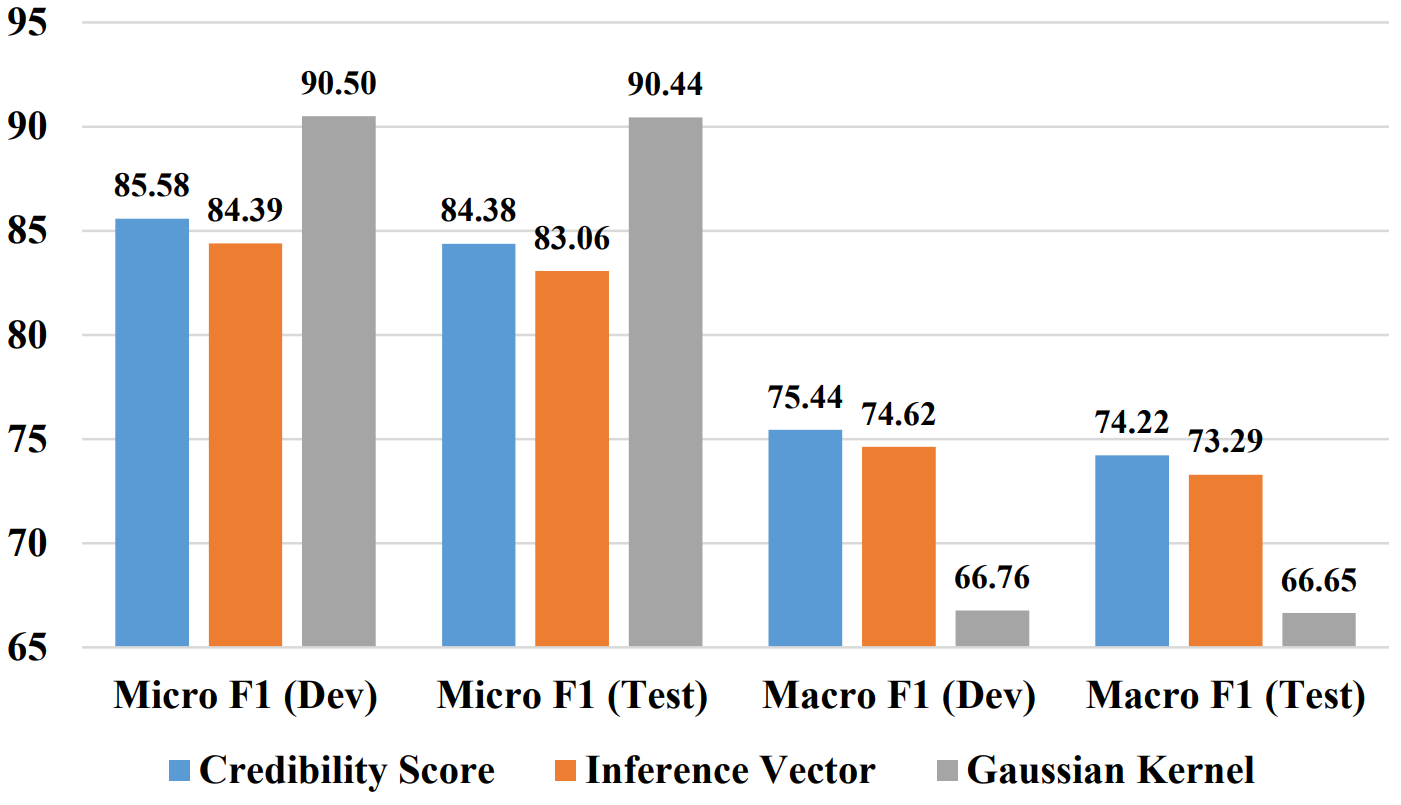}}
\caption{D{\small OC}NLI performance with different fusion methods.}
\label{fusion method}
\end{figure}

Performance of different fusion methods are compared in Figure~\ref{fusion method}.
Gaussian kernel pooling tends to incorrectly predicts all samples as \textit{not entailment} and can hardly recognize \textit{entailment} samples.
Thus it comes with much higher micro F1 score but far lower macro F1 score under the pretty unbalanced label distribution (in Table~\ref{dataset}).
This mean pooling fusion method seems not feasible for the task.
On both sets, credibility score minimum pooling outperforms all other fusion methods.
This fusion method also comes with clear logic basis to expand the interpretability of the framework.

\subsection{Interpretability Study}
\label{interpretability study}

\begin{figure*}[!h]
\centerline{\includegraphics[scale=0.55]{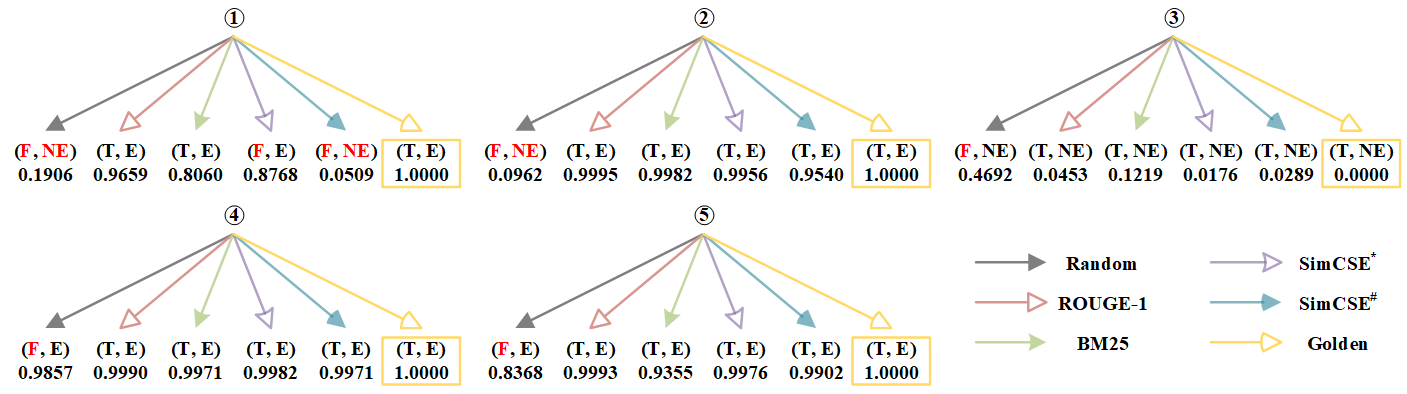}}
\caption{Case study on the sample in Figure~\ref{sample}. For each hypothesis sentence, the elements in the bracket denote whether at least one complete evidence group is obtained,  \textit{T} for \textit{True} while \textit{F} for \textit{False}, and entailment label prediction result, \textit{E} for \textit{entailment} while \textit{NE} for \textit{not entailment}. Wrong predictions are in red. The number below is credibility score. The \textit{Golden} branch denotes the groundtruth.}
\label{explanation}
\end{figure*}

To investigate the interpretability of our framework, we conduct sentence-level evaluation on the subset annotated by ourselves.
As shown in Table~\ref{sentence-level evaluation}, we focus on evidence retrieval and entailment label prediction for hypothesis sentences.

$\bullet$~\textbf{Evidence Retrieval}~For these longer samples (all over 800 words), the random baseline can hardly obtain the evidence, with extremely low evidence recall.
All other retrieval methods can find at least one complete evidence group for most of the hypothesis sentences, with relatively high evidence recall around 85\%.
However, with pretty low evidence precision and F1 score, all these methods will introduce plenty of uninformative sentences as noise.
Thus, the evidence retrieval component may need further improvement.

\begin{table}[!h]
\setlength\tabcolsep{1pt}
\begin{tabular}{ccccccc}
\hline
\multirow{2}{*}{\textbf{\begin{tabular}[c]{@{}c@{}}Retrieval\\ Method\end{tabular}}} & \multicolumn{3}{c}{\textbf{Evi. Retri.}} & \multicolumn{2}{c}{\textbf{Label Pre.}} & \multirow{2}{*}{\textbf{FA}} \\
                                                                                     & \textbf{P}  & \textbf{R}  & \textbf{F1}  & \textbf{Mi F1}      & \textbf{Ma F1}      &                              \\ \hline
Random                                                                               & 6.83        & 17.21       & 9.78         & 66.94               & 63.32               & 13.11                        \\
ROUGE-1                                                                              & \textbf{31.53}       & \textbf{87.43}       & \textbf{46.35}        & 76.23               & 73.92               & \textbf{68.85}                        \\
BM25                                                                                 & 31.53       & 87.16       & 46.31        & 73.22               & 70.46               & 66.12                        \\
SimCSE$^*$                                                                           & 30.38       & 84.97       & 44.76        & \textbf{76.50}               & \textbf{74.69}               & 67.49                        \\
SimCSE$^\#$                                                                          & 31.31       & 84.97       & 45.76        & 73.50               & 70.48               & 65.03                        \\ \hline
\end{tabular}
\caption{Sentence-level evaluation results. \textit{Evi. Retri.}, \textit{Label Pre.}, and \textit{FA} respectively denote \textit{Evidence Retrieval}, \textit{Label Prediction} and \textit{Full Accuracy}.}
\label{sentence-level evaluation}
\end{table}

$\bullet$~\textbf{Entailment Label Prediction}~Regardless of poor retrieval performance, the random baseline still shows relatively high performance on entailment label prediction.
This is due to \textit{evidence dependency bias}.
\textit{Entailment} samples strictly request at least one complete evidence group.
However, \textit{not entailment} samples are insensitive to evidence retrieval.
With complete evidence group obtained, they are taken as \textit{conflict} to the premise, while with evidence missing, they will be taken as \textit{not mentioned}.
Both situations are considered as \textit{not entailment}.
This kind of bias will also contribute to the high document-level performance of the random baseline in Table~\ref{retrieval}.
All other retrieval methods obtain much higher performance with the power of evidence.
However, it seems that high evidence recall is not promising for more accurate entailment label prediction for hypothesis sentences.
This may be also due to evidence dependency bias.
Besides, with evidence precision at only around 30\%, evidence noise is also an important issue, the influence of which is difficult to estimate.
Moreover, the high full accuracy score means for more than 65\% hypothesis sentences, our framework can find at least one complete evidence group and correctly predict their entailment label.
Therefore, taking that sentence-level annotation is not available during training, our R$^2$F framework is able to give more interpretable prediction results and help to locate the semantic inconsistency.

\subsection{Case Study}

To further display the interpretability of our framework, we conduct case study on the sample in Figure~\ref{sample}.
As shown in Figure~\ref{explanation}, for this long sample (about 667 words in total), the random baseline is pretty weak and cannot find complete evidence group for any hypothesis sentence, although it can obtain correct prediction for the sample.
This also demonstrates the importance of evidence discovery for interpretability study.
By contrast, all other retrieval methods can obtain complete evidence groups, and contribute to correctly predict sentence-level entailment label for most of the hypothesis sentences.
Thus the framework can give more interpretable prediction results, and accurately locate the semantic inconsistency in the third hypothesis sentence.
Furthermore, with the two sparse retrieval methods, our framework can even successfully  handle all the hypothesis sentences.
However, the two dense methods fail to obtain complete evidence group for the first hypothesis sentence.
As shown in Figure~\ref{sample}, this one requests two evidences, whose most words are related to the first evidence while only a few words are related to the second one.
Moreover, it rephrases the second one with totally different expressions.
Therefore, it is difficult to find the second one.
This may be the exact situation that leads to the lower evidence recall of dense retrieval methods than sparse ones in Table~\ref{sentence-level evaluation}.

\section{Related Work}

Natural language inference is a fundamental yet important task in natural language processing. 
For sentence-level NLI, several benchmarks~\cite{bowman:snli,samuel:mnli,wang:qnli,nie:anli} have been proposed and they have been attracting research attention.
Moreover, \citet{yuta:contract} propose Contract NLI targeting the legal and business domain, which is a small scale benchmark.
The premises are long contract documents while the hypotheses are actually single sentences.
Besides, \citet{tian:cite} suggest to debias natural language inference and understanding models through causal intervention and counterfactual reasoning.
\citet{li:cite} adopt commonsense inference to enhance future event generation.
For document-level NLI, \citet{yin:docnli} propose the first large scale benchmark D{\small OC}NLI based on a set of early benchmarks. 
Current models for the task still largely follow sentence-level NLI.
Differently, in this paper, we emphasize the importance of evidence discovery and aim at a general solution for the task.

\section{Conclusion}

In this paper, we propose R$^2$F framework as a general solution for D{\small OC}NLI task and contribute complementary annotation on D{\small OC}NLI dataset.
Our experimental results show that our framework can obtain state-of-the-art performance.
Besides, the framework is robust for diverse retrieval methods, and consistently obtains higher performance on samples with varying length, especially longer samples.
Moreover, the framework can give more interpretable prediction results and help to locate the semantic inconsistency.
In the future, we will explore an end-to-end framework for the task.

\section*{Limitations}

The main limitation of our R$^2$F framework is that the framework is a \textit{pipeline} one rather than an \textit{end-to-end} one.
For efficiency, the evidence retrieval component is an independent unit, which provides evidence input for the jointly trained reading and fusion components.
However, the evidence retrieval component in our pipeline framework will not bring additional heavy computation.
First, it can be conducted offline efficiently. 
The sparse retrieval methods only involve item frequency counting, and the dense ones are based on pretrained sentence embeddings without further fine-tuning.
Second, there are many acceleration techniques for retrieval in industrial field.

Furthermore, we will try to improve the evidence retrieval component in the future. 
On one hand, we will try to utilize document-level entailment label to improve sentence-level evidence retrieval. 
On the other hand, we will also explore an efficient end-to-end model, which may benefit from reinforcement learning~\cite{reinforce1,lei:reinforcement}, reparameterization trick~\cite{chris:repara,eric:repara}, or Expectation-Maximum algorithm~\cite{em}.
However, the evidence dependency bias issue, discussed in Section~\ref{interpretability study}, will pose a great challenge. 
That is, \textit{entailment} samples strictly request complete evidence groups, while \textit{not entailment} samples are insensitive to evidence retrieval. 

\section*{Acknowledgments}
This research is supported by the Singapore Ministry of Education (MOE) Academic Research Fund (AcRF) Tier 1 grant, under the RIE2020 Industry Alignment Fund – Industry Collaboration Projects (IAF-ICP) Funding Initiative, as well as cash and in-kind contribution from the industry partner(s).

\bibliography{doc_nli}

\begin{thebibliography}{41}
\expandafter\ifx\csname natexlab\endcsname\relax\def\natexlab#1{#1}\fi

\bibitem[{Arora et~al.(2022)Arora, Asri, Bahuleyan, and Cheung}]{arora:bias}
Kushal Arora, Layla~El Asri, Hareesh Bahuleyan, and Jackie Chi~Kit Cheung.
  2022.
\newblock \href {https://aclanthology.org/2022.findings-acl.58} {Why exposure
  bias matters: An imitation learning perspective of error accumulation in
  language generation}.
\newblock In \emph{Findings of 2022 the Association for Computational
  Linguistics}, pages 700--710. Association for Computational Linguistics.

\bibitem[{Beltagy et~al.(2020)Beltagy, Peters, and Cohan}]{iz:longformer}
Iz~Beltagy, Matthew~E. Peters, and Arman Cohan. 2020.
\newblock \href {https://arxiv.org/abs/2004.05150} {Longformer: The
  long-document transformer}.
\newblock \emph{CoRR}, abs/2004.05150.

\bibitem[{Bengio et~al.(2015)Bengio, Vinyals, Jaitly, and
  Shazeer}]{bengio:bias}
Samy Bengio, Oriol Vinyals, Navdeep Jaitly, and Noam Shazeer. 2015.
\newblock \href
  {https://proceedings.neurips.cc/paper/2015/hash/e995f98d56967d946471af29d7bf99f1-Abstract.html}
  {Scheduled sampling for sequence prediction with recurrent neural networks}.
\newblock In \emph{Proceedings of the 2015 Advances in Neural Information
  Processing System}, pages 1171--1179.

\bibitem[{Bowman et~al.(2015)Bowman, Angeli, Potts, and Manning}]{bowman:snli}
Samuel~R. Bowman, Gabor Angeli, Christopher Potts, and Christopher~D. Manning.
  2015.
\newblock \href {https://doi.org/10.18653/v1/d15-1075} {A large annotated
  corpus for learning natural language inference}.
\newblock In \emph{Proceedings of the 2015 Conference on Empirical Methods in
  Natural Language Processing}, pages 632--642. The Association for
  Computational Linguistics.

\bibitem[{Chen et~al.(2017)Chen, Zhu, Ling, Wei, Jiang, and Inkpen}]{esim}
Qian Chen, Xiaodan Zhu, Zhen{-}Hua Ling, Si~Wei, Hui Jiang, and Diana Inkpen.
  2017.
\newblock \href {https://doi.org/10.18653/v1/P17-1152} {Enhanced lstm for
  natural language inference}.
\newblock In \emph{Proceedings of the 55th Annual Meeting of the Association
  for Computational Linguistics}, pages 1657--1668. Association for
  Computational Linguistics.

\bibitem[{Dai et~al.(2019)Dai, Yang, Yang, Carbonell, Le, and
  Salakhutdinov}]{dai:xl}
Zihang Dai, Zhilin Yang, Yiming Yang, Jaime~G. Carbonell, Quoc~Viet Le, and
  Ruslan Salakhutdinov. 2019.
\newblock \href {https://doi.org/10.18653/v1/p19-1285} {Transformer-xl:
  Attentive language models beyond a fixed-length context}.
\newblock In \emph{Proceedings of the 57th Conference of the Association for
  Computational Linguistics}, pages 2978--2988. Association for Computational
  Linguistics.

\bibitem[{Dempster et~al.(1977)Dempster, Laird, and Rubin}]{em}
A.~P. Dempster, N.~M. Laird, and D.~B. Rubin. 1977.
\newblock \href
  {https://rss.onlinelibrary.wiley.com/doi/abs/10.1111/j.2517-6161.1977.tb01600.x}
  {Maximum likelihood from incomplete data via the em algorithm}.
\newblock \emph{Journal of the Royal Statistical Society: Series B
  (Methodological)}, 39(1):1--22.

\bibitem[{Ding et~al.(2020)Ding, Zhou, Yang, and Tang}]{ding:cogltx}
Ming Ding, Chang Zhou, Hongxia Yang, and Jie Tang. 2020.
\newblock \href
  {https://proceedings.neurips.cc/paper/2020/hash/96671501524948bc3937b4b30d0e57b9-Abstract.html}
  {Cogltx: Applying {BERT} to long texts}.
\newblock In \emph{Proceedings of the 2020 Annual Conference on Neural
  Information Processing Systems}.

\bibitem[{Gao et~al.(2021)Gao, Yao, and Chen}]{gao:simcse}
Tianyu Gao, Xingcheng Yao, and Danqi Chen. 2021.
\newblock \href {https://doi.org/10.18653/v1/2021.emnlp-main.552} {Simcse:
  Simple contrastive learning of sentence embeddings}.
\newblock In \emph{Proceedings of the 2021 Conference on Empirical Methods in
  Natural Language Processing}, pages 6894--6910. Association for Computational
  Linguistics.

\bibitem[{Huang et~al.(2022)Huang, McKeown, Nakov, Choi, and Ji}]{ji:real}
Kung~Hsiang Huang, Kathleen~R. McKeown, Preslav Nakov, Yejin Choi, and Heng Ji.
  2022.
\newblock \href {https://doi.org/10.48550/arXiv.2203.05386} {Faking fake news
  for real fake news detection: Propaganda-loaded training data generation}.
\newblock \emph{CoRR}, abs/2203.05386.

\bibitem[{Jang et~al.(2017)Jang, Gu, and Poole}]{eric:repara}
Eric Jang, Shixiang Gu, and Ben Poole. 2017.
\newblock \href {https://openreview.net/forum?id=rkE3y85ee} {Categorical
  reparameterization with gumbel-softmax}.
\newblock In \emph{5th International Conference on Learning Representations}.
  OpenReview.net.

\bibitem[{Jawahar et~al.(2022)Jawahar, Abdul{-}Mageed, and
  Lakshmanan}]{laks:detail}
Ganesh Jawahar, Muhammad Abdul{-}Mageed, and Laks V.~S. Lakshmanan. 2022.
\newblock \href {https://aclanthology.org/2022.acl-short.10} {Automatic
  detection of entity-manipulated text using factual knowledge}.
\newblock In \emph{Proceedings of the 60th Annual Meeting of the Association
  for Computational Linguistics}, pages 86--93. Association for Computational
  Linguistics.

\bibitem[{Kipf and Welling(2017)}]{kipf:gcn}
Thomas~N. Kipf and Max Welling. 2017.
\newblock \href {https://openreview.net/forum?id=SJU4ayYgl} {Semi-supervised
  classification with graph convolutional networks}.
\newblock In \emph{Proceedings of 5th International Conference on Learning
  Representations}. OpenReview.net.

\bibitem[{Kitaev et~al.(2022)Kitaev, Lu, and Klein}]{lu:parsing}
Nikita Kitaev, Thomas Lu, and Dan Klein. 2022.
\newblock \href {https://aclanthology.org/2022.acl-long.220} {Learned
  incremental representations for parsing}.
\newblock In \emph{Proceedings of the 60th Annual Meeting of the Association
  for Computational Linguistics}, pages 3086--3095. Association for
  Computational Linguistics.

\bibitem[{Koreeda and Manning(2021)}]{yuta:contract}
Yuta Koreeda and Christopher~D. Manning. 2021.
\newblock \href {https://doi.org/10.18653/v1/2021.findings-emnlp.164}
  {Contractnli: {A} dataset for document-level natural language inference for
  contracts}.
\newblock In \emph{Findings of the 2021 Conference on Empirical Methods in
  Natural Language Processing}, pages 1907--1919. Association for Computational
  Linguistics.

\bibitem[{Lei et~al.(2016)Lei, Barzilay, and Jaakkola}]{lei:reinforcement}
Tao Lei, Regina Barzilay, and Tommi~S. Jaakkola. 2016.
\newblock \href {https://doi.org/10.18653/v1/d16-1011} {Rationalizing neural
  predictions}.
\newblock In \emph{Proceedings of the 2016 Conference on Empirical Methods in
  Natural Language Processing}, pages 107--117. The Association for
  Computational Linguistics.

\bibitem[{Lin et~al.(2022)Lin, Cao, Huang, Li, Hu, Wen, and Wang}]{li:cite}
Li~Lin, Yixin Cao, Lifu Huang, Shuang Li, Xuming Hu, Lijie Wen, and Jianmin
  Wang. 2022.
\newblock \href {https://doi.org/10.1145/3477495.3532080} {What makes the story
  forward?: Inferring commonsense explanations as prompts for future event
  generation}.
\newblock In \emph{Proceedings of the 45th International ACM SIGIR Conference
  on Research and Development in Information Retrieval}, pages 1098--1109. ACM.

\bibitem[{Liu et~al.(2019)Liu, Ott, Goyal, Du, Joshi, Chen, Levy, Lewis,
  Zettlemoyer, and Stoyanov}]{liu:roberta}
Yinhan Liu, Myle Ott, Naman Goyal, Jingfei Du, Mandar Joshi, Danqi Chen, Omer
  Levy, Mike Lewis, Luke Zettlemoyer, and Veselin Stoyanov. 2019.
\newblock \href {http://arxiv.org/abs/1907.11692} {Roberta: A robustly
  optimized bert pretraining approach}.
\newblock \emph{CoRR}, abs/1907.11692.

\bibitem[{Liu et~al.(2020)Liu, Xiong, Sun, and Liu}]{liu:kernel}
Zhenghao Liu, Chenyan Xiong, Maosong Sun, and Zhiyuan Liu. 2020.
\newblock \href {https://doi.org/10.18653/v1/2020.acl-main.655} {Fine-grained
  fact verification with kernel graph attention network}.
\newblock In \emph{Proceedings of the 58th Annual Meeting of the Association
  for Computational Linguistics}, pages 7342--7351. Association for
  Computational Linguistics.

\bibitem[{Maddison et~al.(2017)Maddison, Mnih, and Teh}]{chris:repara}
Chris~J. Maddison, Andriy Mnih, and Yee~Whye Teh. 2017.
\newblock \href {https://openreview.net/forum?id=S1jE5L5gl} {The concrete
  distribution: A continuous relaxation of discrete random variables}.
\newblock In \emph{5th International Conference on Learning Representations}.
  OpenReview.net.

\bibitem[{Mao et~al.(2022)Mao, Wu, Ni, Zhang, Zhang, Yu, Deb, Zhu, Awadallah,
  and Radev}]{mao:rouge1}
Ziming Mao, Chen~Henry Wu, Ansong Ni, Yusen Zhang, Rui Zhang, Tao Yu,
  Budhaditya Deb, Chenguang Zhu, Ahmed~Hassan Awadallah, and Dragomir~R. Radev.
  2022.
\newblock \href {https://aclanthology.org/2022.acl-long.118} {Dyle: Dynamic
  latent extraction for abstractive long-input summarization}.
\newblock In \emph{Proceedings of the 60th Annual Meeting of the Association
  for Computational Linguistics}, pages 1687--1698. Association for
  Computational Linguistics.

\bibitem[{Narayan et~al.(2018)Narayan, Cohen, and Lapata}]{shay:summary}
Shashi Narayan, Shay~B. Cohen, and Mirella Lapata. 2018.
\newblock \href {https://doi.org/10.18653/v1/d18-1206} {Don't give me the
  details, just the summary! topic-aware convolutional neural networks for
  extreme summarization}.
\newblock In \emph{Proceedings of the 2018 Conference on Empirical Methods in
  Natural Language Processing}, pages 1797--1807. Association for Computational
  Linguistics.

\bibitem[{Nie et~al.(2020)Nie, Williams, Dinan, Bansal, Weston, and
  Kiela}]{nie:anli}
Yixin Nie, Adina Williams, Emily Dinan, Mohit Bansal, Jason Weston, and Douwe
  Kiela. 2020.
\newblock \href {https://doi.org/10.18653/v1/2020.acl-main.441} {Adversarial
  nli: A new benchmark for natural language understanding}.
\newblock In \emph{Proceedings of the 58th Annual Meeting of the Association
  for Computational Linguistics}, pages 4885--4901. Association for
  Computational Linguistics.

\bibitem[{Park et~al.(2022)Park, Vyas, and Shah}]{park}
Hyunji~Hayley Park, Yogarshi Vyas, and Kashif Shah. 2022.
\newblock \href {https://aclanthology.org/2022.acl-short.79} {Efficient
  classification of long documents using transformers}.
\newblock In \emph{Proceedings of the 60th Annual Meeting of the Association
  for Computational Linguistics}, pages 702--709. Association for Computational
  Linguistics.

\bibitem[{Sandhaus(2008)}]{evan:summary}
Evan Sandhaus. 2008.
\newblock \href {https://catalog.ldc.upenn.edu/LDC2008T19} {The new york times
  annotated corpus}.
\newblock In \emph{LDC corpora}. Linguistic Data Consortium.

\bibitem[{Sheng et~al.(2022)Sheng, Cao, Zhang, Li, Wang, and Zhu}]{sheng:zoom}
Qiang Sheng, Juan Cao, Xueyao Zhang, Rundong Li, Danding Wang, and Yongchun
  Zhu. 2022.
\newblock \href {https://aclanthology.org/2022.acl-long.311} {Zoom out and
  observe: News environment perception for fake news detection}.
\newblock In \emph{Proceedings of the 60th Annual Meeting of the Association
  for Computational Linguistics}, pages 4543--4556. Association for
  Computational Linguistics.

\bibitem[{Thorne et~al.(2018)Thorne, Vlachos, Christodoulopoulos, and
  Mittal}]{thorne:fever}
James Thorne, Andreas Vlachos, Christos Christodoulopoulos, and Arpit Mittal.
  2018.
\newblock \href {https://doi.org/10.18653/v1/n18-1074} {Fever: a large-scale
  dataset for fact extraction and verification}.
\newblock In \emph{Proceedings of the 2018 Conference of the North American
  Chapter of the Association for Computational Linguistics:Human Language
  Technologies}, pages 809--819. Association for Computational Linguistics.

\bibitem[{Tian et~al.(2022)Tian, Cao, Zhang, and Xing}]{tian:cite}
Bing Tian, Yixin Cao, Yong Zhang, and Chunxiao Xing. 2022.
\newblock \href {https://ojs.aaai.org/index.php/AAAI/article/view/21389}
  {Debiasing nlu models via causal intervention and counterfactual reasoning}.
\newblock In \emph{Proceedings of the 36th AAAI Conference on Artificial
  Intelligence}, pages 11376--11384. AAAI Press.

\bibitem[{Vaswani et~al.(2017)Vaswani, Shazeer, Parmar, Uszkoreit, Jones,
  Gomez, Kaiser, and Polosukhin}]{ashish:attention}
Ashish Vaswani, Noam Shazeer, Niki Parmar, Jakob Uszkoreit, Llion Jones,
  Aidan~N. Gomez, Lukasz Kaiser, and Illia Polosukhin. 2017.
\newblock \href
  {https://proceedings.neurips.cc/paper/2017/hash/3f5ee243547dee91fbd053c1c4a845aa-Abstract.html}
  {Attention is all you need}.
\newblock In \emph{Proceedings of the 2017 Advances in Neural Information
  Processing Systems}, pages 5998--6008.

\bibitem[{Velickovic et~al.(2018)Velickovic, Cucurull, Casanova, Romero, Lio,
  and Bengio}]{petar:gat}
Petar Velickovic, Guillem Cucurull, Arantxa Casanova, Adriana Romero, Pietro
  Lio, and Yoshua Bengio. 2018.
\newblock \href {https://openreview.net/forum?id=rJXMpikCZ} {Graph attention
  networks}.
\newblock In \emph{Proceedings of 6th International Conference on Learning
  Representations}. OpenReview.net.

\bibitem[{Wang et~al.(2019)Wang, Singh, Michael, Hill, Levy, and
  Bowman}]{wang:qnli}
Alex Wang, Amanpreet Singh, Julian Michael, Felix Hill, Omer Levy, and
  Samuel~R. Bowman. 2019.
\newblock \href {https://openreview.net/forum?id=rJ4km2R5t7} {Glue: A
  multi-task benchmark and analysis platform for natural language
  understanding}.
\newblock In \emph{Proceedings of the7th International Conference on Learning
  Representations}. OpenReview.net.

\bibitem[{Wang et~al.(2021)Wang, Tang, Duan, Wei, Huang, Ji, Cao, Jiang, and
  Zhou}]{wang:kadapter}
Ruize Wang, Duyu Tang, Nan Duan, Zhongyu Wei, Xuanjing Huang, Jianshu Ji,
  Guihong Cao, Daxin Jiang, and Ming Zhou. 2021.
\newblock \href {https://doi.org/10.18653/v1/2021.findings-acl.121} {K-adapter:
  Infusing knowledge into pre-trained models with adapters}.
\newblock In \emph{Findings of the 2021 Association for Computational
  Linguistics}, pages 1405--1418. Association for Computational Linguistics.

\bibitem[{Williams et~al.(2018)Williams, Nangia, and Bowman}]{samuel:mnli}
Adina Williams, Nikita Nangia, and Samuel~R. Bowman. 2018.
\newblock \href {https://doi.org/10.18653/v1/n18-1101} {A broad-coverage
  challenge corpus for sentence understanding through inference}.
\newblock In \emph{Proceedings of the 2018 Conference of the North American
  Chapter of the Association for Computational Linguistics: Human Language
  Technologies,}, pages 1112--1122. Association for Computational Linguistics.

\bibitem[{Williams(1992)}]{reinforce1}
Ronald~J. Williams. 1992.
\newblock \href {https://doi.org/10.1007/BF00992696} {Simple statistical
  gradient-following algorithms for connectionist reinforcement learning}.
\newblock \emph{Machine Learning}, 8:229--256.

\bibitem[{Wolpert and Macready(1997)}]{NFL}
David~H. Wolpert and William~G. Macready. 1997.
\newblock \href {https://doi.org/10.1109/4235.585893} {No free lunch theorems
  for optimization}.
\newblock \emph{IEEE Transactions on Evolutionary Computation}, 1(1):67--82.

\bibitem[{Wu et~al.(2021)Wu, Wu, Qi, and Huang}]{wu:hi}
Chuhan Wu, Fangzhao Wu, Tao Qi, and Yongfeng Huang. 2021.
\newblock \href {https://doi.org/10.18653/v1/2021.acl-short.107}
  {Hi-transformer: Hierarchical interactive transformer for efficient and
  effective long document modeling}.
\newblock In \emph{Proceedings of the 59th Annual Meeting of the Association
  for Computational Linguistics}, pages 848--853. Association for Computational
  Linguistics.

\bibitem[{Xiong et~al.(2017)Xiong, Dai, Callan, Liu, and Power}]{xiong:kernel}
Chenyan Xiong, Zhuyun Dai, Jamie Callan, Zhiyuan Liu, and Russell Power. 2017.
\newblock \href {https://doi.org/10.1145/3077136.3080809} {End-to-end neural
  ad-hoc ranking with kernel pooling}.
\newblock In \emph{Proceedings of the 40th International ACM SIGIR Conference
  on Research and Development in Information Retrieval}, pages 55--64. ACM.

\bibitem[{Yin et~al.(2021)Yin, Radev, and Xiong}]{yin:docnli}
Wenpeng Yin, Dragomir~R. Radev, and Caiming Xiong. 2021.
\newblock \href {https://doi.org/10.18653/v1/2021.findings-acl.435} {Docnli: A
  large-scale dataset for document-level natural language inference}.
\newblock In \emph{Findings of the 2021 Association for Computational
  Linguistic}, pages 4913--4922. Association for Computational Linguistics.

\bibitem[{Zhang et~al.(2022)Zhang, Ni, Mao, Wu, Zhu, Deb, Awadallah, Radev, and
  Zhang}]{zhang:rouge2}
Yusen Zhang, Ansong Ni, Ziming Mao, Chen~Henry Wu, Chenguang Zhu, Budhaditya
  Deb, Ahmed~Hassan Awadallah, Dragomir~R. Radev, and Rui Zhang. 2022.
\newblock \href {https://aclanthology.org/2022.acl-long.112} {$summ^n$: A
  multi-stage summarization framework for long input dialogues and documents}.
\newblock In \emph{Proceedings of the 60th Annual Meeting of the Association
  for Computational Linguistics}, pages 1592--1604. Association for
  Computational Linguistics.

\bibitem[{Zhang et~al.(2019)Zhang, Han, Liu, Jiang, Sun, and Liu}]{zhang:erine}
Zhengyan Zhang, Xu~Han, Zhiyuan Liu, Xin Jiang, Maosong Sun, and Qun Liu. 2019.
\newblock \href {https://doi.org/10.18653/v1/p19-1139} {Ernie: Enhanced
  language representation with informative entities}.
\newblock In \emph{Proceedings of the 57th Conference of the Association for
  Computational Linguistics}, pages 1441--1451. Association for Computational
  Linguistics.

\bibitem[{Zhong et~al.(2020)Zhong, Liu, Chen, Wang, Qiu, and Huang}]{qiu:match}
Ming Zhong, Pengfei Liu, Yiran Chen, Danqing Wang, Xipeng Qiu, and Xuanjing
  Huang. 2020.
\newblock \href {https://doi.org/10.18653/v1/2020.acl-main.552} {Extractive
  summarization as text matching}.
\newblock In \emph{Proceedings of the 58th Annual Meeting of the Association
  for Computational Linguistics}, pages 6197--6208. Association for
  Computational Linguistics.

\end{thebibliography}

\appendix

\section{Statistic~Information~of~Complementary~Annotation }
\label{annotation example}

\begin{table*}[h]
\centering
\setlength\tabcolsep{4pt}
\begin{tabular}{ccccccccc}
\hline
\multirow{2}{*}{\textbf{Model}} & \multicolumn{3}{c}{\textbf{Entailment}}          & \multicolumn{3}{c}{\textbf{Not\_Entailment}}     & \multicolumn{2}{c}{\textbf{Total}}    \\
                                & \textbf{P}     & \textbf{R}     & \textbf{F1}    & \textbf{P}     & \textbf{R}     & \textbf{F1}    & \textbf{Micro F1} & \textbf{Macro F1} \\ \hline
Concatenation-Longformer$_{base}\clubsuit$         & -              & -              & 46.18          & -              & -              & -              & -                 & -                 \\
Concatenation-Longformer$_{base}\spadesuit$        & 31.09          & 81.33          & 44.99          & 96.60          & 74.64          & 84.22          & 75.47             & 64.60             \\
Concatenation-RoBERTa$_{base}\spadesuit$           & 42.27          & \textbf{90.82} & 57.69          & \textbf{98.45} & 82.55          & 89.81          & 83.57             & 73.75             \\
Semantic Match-RoBERTa$_{base}\triangle$           & 31.01          & 84.32          & 45.34          & 97.09          & 73.61          & 83.74          & 74.93             & 64.54             \\
\textit{ours} $\rm R^2F$-RoBERTa$_{base}$          & \textbf{45.54} & 86.43          & \textbf{59.66} & 97.82          & \textbf{85.46} & \textbf{91.22} & \textbf{85.58}    & \textbf{75.44}    \\ \hline \hline
Concatenation-RoBERTa$_{large}\clubsuit$           & 47.15          & \textbf{95.09} & 63.04          & \textbf{99.19} & 85.01          & 91.55          & 86.25             & 77.30             \\
Concatenation-RoBERTa$_{large}\spadesuit$          & 47.04          & 94.82          & 62.89          & 99.15          & 84.99          & 91.52          & 86.20             & 77.21             \\
\textit{ours} $\rm R^2F$-RoBERTa$_{large}$         & \textbf{53.70} & 87.10          & \textbf{66.43} & 98.01          & \textbf{89.43} & \textbf{93.53} & \textbf{89.15}    & \textbf{79.98}    \\ \hline
\end{tabular}
\caption{Model performance on the \textit{dev} set. $\clubsuit$ and $\spadesuit$ denote the \textit{original} and \textit{reproduced} results of the model from \citet{yin:docnli}. $\triangle$ denotes the \textit{reproduced} results of the model modified from \citet{qiu:match}.}
\label{dev results}
\end{table*}

Due to  heavy workload and great complexity, we manually annotate 100 longer samples (all over 800 words) randomly selected from the \textit{test} set, which contain more than 350 hypothesis sentences in total.
Among the samples, 84\% are annotated as \textit{not entailment}, while 16\% are annotated as \textit{entailment}.
However, among the hypothesis sentences, about 43\% are annotated as \textit{not entailment}, while about 57\% are annotated as \textit{entailment}.
The great label distribution difference is natural.
For \textit{entailment} samples, all hypothesis sentences are entailed by the premise.
However, \textit{not entailment} samples can still contain entailed  hypothesis sentences.
Besides, about 47\% hypothesis sentences involve cross-sentence inference and request multiple evidence sentences.
Moreover, about 17\% hypothesis sentences are annotated with more than one evidence groups.
These also indicate the great complexity of fine-grained sentence-level annotation.

\section{Detailed Experiment Setup}
\label{setup}

Our $\rm{R^2F}$ framework is implemented through Pytorch 1.8.0 and hugging face transformers\footnote{\url{https://huggingface.co/}}. 
All experiments are conducted on a computation node with Nvidia 40G A100 GPUs. 
For evidence retrieval, we set \textit{K} as 5 to remain top 5 sentences as evidences during retrieval.
We adopt RoBERTa~\cite{liu:roberta} as encoder, including base and large version.
For all experiments, we adopt AdamW optimizer, keep a random number seed of 42, set max input length as 256, and set mini batch size as 8 with gradient accumulation step as 4.
For base encoder, we choose initial learning rate as 1e-5, while for large encoder, we choose initial learning rate as 5e-6.
We train 5 epochs, evaluate each 3750 steps, and choose the model parameters with the highest performance on the development set.
For Gaussian Kernel Pooling, we keep 11 kernels with the same width of 0.01.
However, the mean values of the kernels come from a uniform distribution within the interval of [0.0, 1.0].
During prediction, we adopt a threshold of 0.5.

\section{Detailed Baseline}
\label{baseline}

Since D{\small OC}NLI is still a new task, we adopt the concatenation model from~\citet{yin:docnli}, and modify the semantic match model from~\citet{qiu:match} for comparison.

$\bullet$~\textbf{Concatenation}~We concatenate the hypothesis and the premise documents into a sequence as input.
Overlong samples will be truncated to max input length.
Then we adopt the hidden state of the special [CLS] token as the sample representation for binary classification.

$\bullet$~\textbf{Semantic Match}~We respectively encode the hypothesis and premise documents to obtain their own document-level vector representation.
For documents exceeding the max input length, we split them into chunks with sliding window.
Then inspired by~\citet{esim}, we enhance the vector representations of these two for classification.

\section{Performance on Development Set}
\label{performance on dev set}

\begin{figure}[!h]
\centering
\centerline{\includegraphics[scale=0.49]{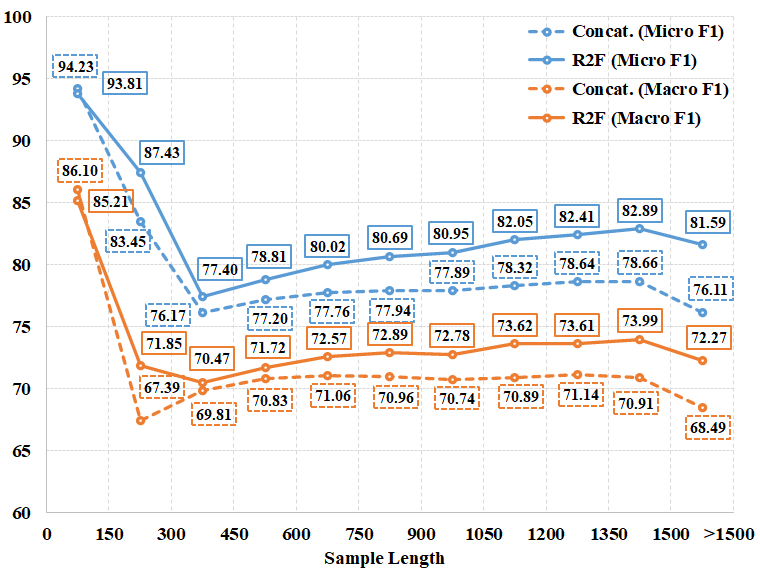}}
\caption{Model performance with varying sample length on the \textit{dev} set. Concatenation baseline and R$^2$F framework with RoBERTa$_{base}$ encoder are compared. The horizontal axis is the sample length in the number of words. The vertical axis is the performance.}
\label{dev length}
\end{figure}

The detailed model performance on the development set are shown in Table \ref{dev results} and Figure \ref{dev length}.
The situations are similar to those on the text set.
Our framework obtains the highest performance on the development set.
Furthermore, except for the slight performance decrease on samples no longer than 150 words, it greatly and consistently outperforms the strongest concatenation baseline on samples with varying length, especially on longer samples.
These also show the strength of our R$^2$F framework on the task.

\section{Influence of \textit{K} Value}

To investigate the influence of evidence retrieval, we also pay attention to  the value of \textit{K}, which is the hyperparameter about how many sentences to remain during retrieval.
As shown in Table \ref{K value}, the framework is pretty sensitive to the value.
Specifically, the highest performance is obtained with \textit{K} as 5.
Besides, with \textit{K} as 3, 6 or 7, the framework can also obtain promoted performance than the concatenation baseline.
However, with \textit{K} as 4, it shows similar performance with the baseline.
As discussed in Section~\ref{evidence retrieval}, \textit{K} is to balance evidence precision and recall.
Lower values pursue higher evidence precision, but may lead to evidence missing, while higher values guarantee higher evidence recall, but may introduce too much noise.
Besides, the choice of the value is also closely related to data distribution.
The high sensitivity to the value indicates that the evidence retrieval process will need further improvement, where a possible way is to utilize the document-level entailment label to improve it.

\begin{table}[!h]
\centering
\begin{tabular}{ccccc}
\hline
\multirow{2}{*}{\textbf{K}} & \multicolumn{2}{c}{\textbf{Dev}} & \multicolumn{2}{c}{\textbf{Test}} \\
                            & \textbf{Mi F1}  & \textbf{Ma F1} & \textbf{Mi F1}  & \textbf{Ma F1}  \\ \hline
3                           & 84.79           & 74.50          & 83.05           & 72.65           \\
4                           & 83.68           & 73.14          & 81.68           & 71.30                \\
5                           & \textbf{85.58 } & \textbf{75.44} & \textbf{84.38}  & \textbf{74.22}  \\
6                           & 84.64           & 74.50          & 83.50           & 73.39           \\
7                           & 84.34           & 73.95          & 82.50           & 72.31                \\ \hline
\end{tabular}
\caption{D{\small OC}NLI performance with varying values of \textit{K}.}
\label{K value}
\end{table}

\section{Related~Work~on~Long~Document~Processing} 

For long document processing, two common methods are to truncate the input sequence to the maximum length, or cut into several independent segments with sliding window.
\citet{dai:xl} propose segment-level recurrence mechanism for information interaction among segments.
\citet{iz:longformer} simplify the self-attention mechanism~\cite{ashish:attention} to reduce memory overhead to handle longer input sequence.
Moreover, \citet{ding:cogltx} suggest to introduce two independent iteratively trained models, respectively for neural evidence retrieval and semantic inference.
\citet{wu:hi}  propose hierarchical interactive transformer structure with stacked transformers respectively for sentence and document encoding.
However, a recent study~\cite{park} on text classification task indicates that none of these models will consistently outperform other models across datasets.
The situation is similar to the famous~\textit{No Free Lunch} principle~\cite{NFL}.
Therefore, long document processing may need further exploration.

\end{document}